\documentclass{article}
\usepackage{epsf}

\usepackage{times}
\usepackage{latexsym}
\usepackage{graphicx}
\usepackage{float}
\usepackage{subfigure}
\usepackage{setspace}
\usepackage{comment}
\usepackage{url}
\usepackage{cite}

\begin{document}

\title{Improving Recurrent Neural Networks For Sequence Labelling}

\author{Marco Dinarelli and Isabelle Tellier\\[18pt]
\small LaTTiCe (UMR 8094), CNRS, ENS Paris, Universit\'e Sorbonne Nouvelle - Paris 3\\
\small PSL Research University, USPC (Universit\'e Sorbonne Paris Cit\'e)\\
\small 1 rue Maurice Arnoux, 92120 Montrouge, France\\
\small \url{marco.dinarelli@ens.fr}, \url{isabelle.tellier@univ-paris3.fr}\\[30pt]
}

\maketitle

\begin{abstract}
In this paper we study different types of Recurrent Neural Networks (RNN) for sequence labeling tasks.
We propose two new variants of RNNs integrating improvements for sequence labeling, 
and we compare them to the more traditional Elman and Jordan RNNs.
We compare all models, either traditional or new, on four distinct tasks of sequence labeling: two on Spoken Language Understanding (ATIS and MEDIA); and two of POS tagging for the French Treebank (FTB) and the Penn Treebank (PTB) corpora.
The results show that our new variants of RNNs are always more effective than the others.
\end{abstract}

\section{Introduction}
\label{sec:Intro}

Recurrent Neural Networks (RNN) \cite{jordan:serial,Elman90findingstructure,Schuster:1997:BRNN} are neural models able to take some context into account in their decision function.
For this reason, they are particularly suitable for several NLP tasks, 
in particular sequential information prediction \cite{Collobert:2008:UAN:1390156.1390177,Collobert:2011:NLP:1953048.2078186,RNNforLU:Interspeech:2013,RNNforSLU:Interspeech:2013,Vukotic.etal_2015}.
In RNNs, the contextual information is provided to the model by a loop connection in the network architecture.
This connection allows to use at the current time step one or more pieces of information predicted at previous time steps.
This architecture seems particularly effective for neural networks since it allows to combine the power of distributional representations (or \textit{embeddings}) with the effectiveness of contextual information.

In the literature about RNNs for NLP, two main variants have been proposed, also called ``simple'' RNNs: 
the Elman \cite{Elman90findingstructure} and the Jordan \cite{jordan:serial} RNN models.
The difference between these models lies in the position of the loop connection giving the \textit{recurrent character} to the network: in the Elman RNN, it is put  
in the hidden layer whereas
in the Jordan RNN 
it connects the output layer to the hidden layer.
In this last case, the recurrent connection allows to use, at the current time step, 
the information predicted at previous time steps.
In the last few years, these two types of RNNs have been very successful for language modeling \cite{RNN:Mikolov:Interspeech:2010,RNNExtensions_Mikolov:ICASSP:2011}, 
and for some sequence labeling tasks \cite{RNNforLU:Interspeech:2013,RNNforSLU:Interspeech:2013,Vukotic.etal_2015,XuetAl:ACL:2015,UnsupervidedPOSTagging:PACLIC:2015}.

The intuition at the origin of this article is that embeddings allow a fine and effective modeling not only of words, but also of \emph{labels} and their dependencies, which are very important for sequence labeling.
In this paper, we define two new variants of RNN to achieve this more effective modeling.

In the first variant, the recurrent connection is between the output and the input layers.
In other words, this variant gives labels predicted at previous positions in a sequence as input to the network. Such contextual information is added to the usual input context made of words, 
and both are used to predict the label at the current position in the sequence.
Moreover we modified the hidden layer activity computation with respect to Elman and Jordan RNN, 
so that to take the different information provided by words and labels into account.
From our intuition, thanks to label embeddings and to features-learning abilities of the hidden layer, this variant models in a more effective way label dependencies.
The second variant we propose combines an Elman RNN and our first variant.
This variant can thus exploit both contextual information provided by the previous states of the hidden layer, and the labels predicted at previous positions of a sequence.

\begin{figure}[t, width=\linewidth]
  \centering
        \footnotesize
  \subfigure[Elman RNN]{
    \includegraphics[height=4.5cm]{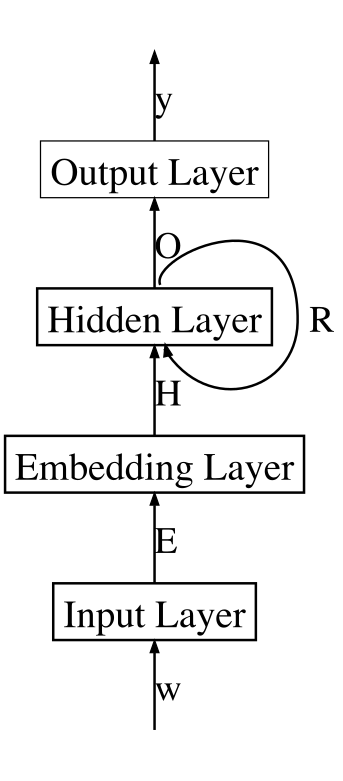}
  }
  \quad
  \subfigure[Jordan RNN]{
    \includegraphics[height=4.5cm]{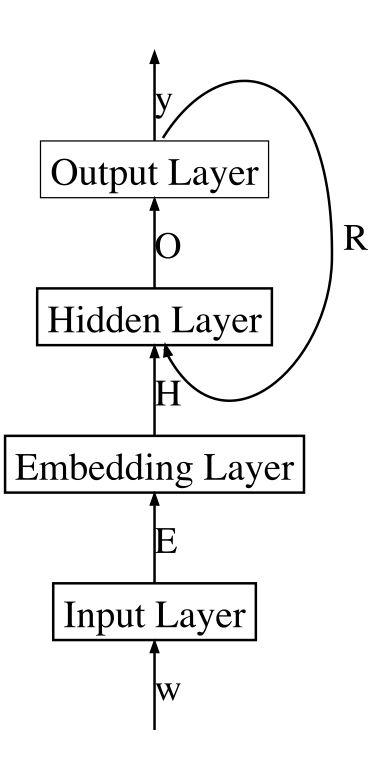}
  }
  \quad
  \subfigure[Our variant of RNN]{
    \includegraphics[height=4.5cm]{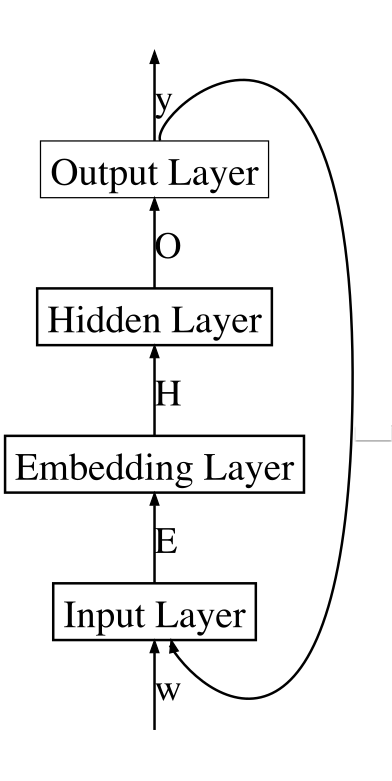}
  }
  \caption{High Level Schema of the Main RNNs Studied in this Work.}
	\label{fig:3architecgtures}
\end{figure}

A high-level schema of the Elman's, Jordan's and our first variant of RNN are shown in Figure~\ref{fig:3architecgtures}. The schema of our second variant can be obtained by adding the recursion of the first one to the Elman architecture. 
In this figure, $w$ is the input word, $y$ is the predicted label, $E$, $H$, $O$ and $R$ are the parameter matrices between each pair of layers: they will be described in details in the next section.
Before, 
it is worth discussing 
the advantages our variants can bring with respect to traditional RNN architectures, and explaining why they are expected to provide better modeling abilities.

First, since the output at previous positions is given as input to the network at the current position,
the contextual information flows across the whole network, affecting each layer's input and output, at both forward and backward phases.
In contrast, in Elman and Jordan RNNs, not all layers are affected at both forward and backward phases.

A second advantage of our variants  is given by label embeddings.
Indeed, the first layer of our RNNs is just a \textit{look-up} table mapping sparse ``one-hot'' representations into distributional representations.\footnote{The ``one-hot'' representation of an element at position $i$ in a dictionary $V$ is a vector of size $|V|$ where the $i$-th component has the value $1$ whereas all the others are $0$.}
Since in our variants the output of the network at previous steps is given as input at the current step, 
the mapping from sparse representations to embeddings involves both words and labels.
Label embeddings can be pre-trained from data as it is usually done for words.
Pre-trained word embeddings, e.g. with \textit{word2vec}, 
have already shown their ability to capture very attractive syntactic and semantic properties \cite{Word2Vec_Mikolov:2013,LinguisticRegularities:Mikolov:2013}.
Using label embeddings, the same properties can be learned also for labels.
More importantly, using several predicted labels as embeddings provide a more effective modeling of label dependencies via the internal state of the network, which is the hidden layer's output.

Another advantage coming from the use of label embeddings and different previous labels as context, 
is an increased robustness of the model to prediction mistakes. This effect comes from the syntactic and semantic properties that embeddings can encode \cite{LinguisticRegularities:Mikolov:2013}.

All these advantages are also supported by our second variant, 
which uses both a label embedding context, like the first variant, 
and the loop connection at the hidden layer, like an Elman RNN.

All RNNs in this article are studied in their 
\textit{forward}, \textit{backward} and \textit{bidirectional} versions \cite{Schuster:1997:BRNN}.
In order to have a fair and straightforward comparison, 
we give the results of our new variants of RNNs together with those obtained with our implementation of Elman and Jordan RNNs.
These implementations are very close to state-of-the-art, 
even if we did not implement every optimization feature.

All models are evaluated on four tasks.
Two are Spoken Language Understanding (SLU) tasks \cite{demori08:SPM}: ATIS \cite{Dahl:1994:ESA:1075812.1075823} and MEDIA \cite{Bonneau-Maynard2006:media}, which can be both modeled as sequence labeling problems.
Two are POS-tagging tasks, one on the French Treebank (FTB)~\cite{FTB:Abeille:2003,Denis:2012:FTB-POStagging} and one on the Penn Treebank (PTB)~\cite{Marcus93buildinga}.
The results we obtain on these tasks with our implementations, 
despite they are not always better than the state-of-the-art, 
provide a stable ranking of different RNN architectures: 
at least one of our variants, most of the time and surprisingly the simpler one, 
is always better than Jordan and Elman RNNs.

In the remainder of the paper, we introduce RNNs and we describe in more details the variants proposed in this work (section~\ref{sec:IRNN}).
In section~\ref{sec:eval}, we describe the corpora used for evaluation and all the results obtained, in comparison with state-of-the-art models.
In section~\ref{sec:conclusions}, we draw our conclusions.

\section{Improving Recurrent Neural Networks}
\label{sec:IRNN}

The RNNs we consider in this work have the same architecture also used for Feedforward Neural Network Language Models (NNLM), described in \cite{Bengio03aneural}.
In this architecture, we have four layers: input, embedding, hidden and output.
Words are given as input to the network as indexes, corresponding to their position in a dictionary $V$.

The index of a word is used to select its embedding (or distributional representation) in a real-valued matrix $ E \in \!R^{|V| X N}$, $|V|$ being the size of the dictionary and $N$ the dimensionality of the embeddings (which is a parameter to be chosen).
We name $E(v(w_t))$ the embedding of the word $w$ given as input at the position $t$ of a sequence.
$v(w_t) = i$ is the index of the word $w_t$ in the dictionary,
and it can be seen alternatively as a ``one-hot'' vector representation (the vector is zero everywhere except at position $v(w)$, where it is $1$).

In contrast to NNLM, RNNs have one more connection, the recursive connection, between two layers, depending on the type of RNNs. As mentioned previously, Elman RNNs have a recursion loop in the hidden layer.
Since this layer encodes the internal representation of the input to the network, 
the recurrent connection of an Elman network allows to keep ``in memory'' words used as input at previous positions in the sequence.
Jordan RNNs have instead a recursion between the output and the hidden layer.
This means that a Jordan RNN can take previous predicted labels into account to predict the label at the current position in a sequence.
For every type of RNN, we call $R$ the matrix of parameters of the recursion connection.

Our implementation of Jordan and Elman RNNs is like in the literature.
\cite{jordan:serial,Elman90findingstructure,Schuster:1997:BRNN}.
\footnote{We also find \cite{RNNforSLU:Interspeech:2013} quite easy to understand, 
even for readers not familiar with RNNs.}

\subsection{RNN Learning}
\label{subsec:learning}

Learning the described RNNs consists in learning the parameters $\Theta = (E, H, O, R)$ between each pair of layers (see Figure~\ref{fig:3architecgtures}), and we omit biases to keep notations lighter.
We use a cross-entropy cost function between the expected label $c_t$ and the predicted label $y_t$ at the position $t$ in the sequence, plus a $L2$ regularization term \cite{PracticalRecommendations:Bengio:2012}:

\begin{equation}
C = - c_t \cdot log( y_t ) + \frac{\lambda}{2} \left | \Theta \right |^2
\end{equation}

$\lambda$ is an hyper-parameter of the model.
Since $y_t$ is a probability distribution over output labels,
we can also view the output of a RNN as the probability of the predicted label $y_t$: $P(y_t | I, \Theta)$.
$I$ is the input given to the network plus the contextual information provided by the recurrent connection.
For Elman RNN $I_{Elman} = w_{t-w} ... w_t ... w_{t+w}, h_{t-1}$, that is the word input context and the output of the hidden layer at the previous position in the sequence.
For Jordan RNN $I_{Jordan} = w_{t-w} ... w_t ... w_{t+w}, y_{t-1}$, that is the same word input context as Elman RNN, and the label predicted at the previous position.
We associate the following decision function to predict the label at position $t$ in a sequence:

\begin{equation}
\label{eq:dec-function}
\tilde{l_t} = argmax_{j \in {1,...,|L|}} P(y_t^j | I, \Theta)
\end{equation}

$\tilde{l_t}$ is a particular discrete label.

We use the back-propagation algorithm and the stochastic gradient descent with momentum \cite{PracticalRecommendations:Bengio:2012} for learning the weights $\Theta$.

\subsection{Learning Variants}
\label{subsec:tricks}

An important choice for learning RNN models concern the back-propagation algorithm.
Indeed, because of the recurrent nature of their architecture, 
in order to properly learn RNNs the Back-Propagation Through Time algorithm (\texttt{BPTT}) should be used \cite{werbos:bptt}.
The BPTT algorithm constists basically in unfolding the recurrent architecture for a choosen number of steps, and then learning the network as a standard Feed-Forward network.
This is supposed to allow RNNs to learn arbitrarily long past context.
However, \cite{RNNExtensions_Mikolov:ICASSP:2011} has shown that RNNs for language modelling learn best with just $5$ time steps in the past.
This may be due to the fact that, at least in NLP tasks, the past information kept ``in memory'' by the network via the recurrent architecture, actually fades away after some time steps.
Moreover, in many NLP tasks, using an arbitrarily-long context on either input or output side, 
doesn't garantee better performances, as increasing the context size also increases the noise.
Since BPTT is quite more expensive than the traditional back-propagation algorithm,
\cite{RNNforSLU:Interspeech:2013} has preferred to use explicit output context in Jordan RNNs and to learn the model with the traditional back-propagation algorithm, not surprisingly without loosing performance.

In this work we use the same variant as \cite{RNNforSLU:Interspeech:2013}.
When using an explicit context of output labels from previous time steps, the hidden layer activity of a Jordan RNN is computed as:

\begin{equation}
\label{eqn:JordanHiddenContextHistory}
h_t = \Sigma( I_t \cdot H + [y_{t-c+1} y_{t-c+2} ... y_{t-1}] \cdot R)
\end{equation}

where $c$ is the size of the history of previous labels that we want to explicitly use as context to predict next label. $[\cdot]$ indicates the concatenation of vectors.

All the modifications applied to the Jordan RNN so far can be applied in a similar way to the Elman RNN.

\subsection{New Variants of RNN}
\label{subsec:NewRNN}

As mentioned in the introduction, the new variants of RNN proposed in this work present two differences with respect to traditional RNNs: i) the recurrent connection is from the output to the input layer, meaning that predicted labels are converted into embeddings in the same way as words; ii) the hidden layer activities are computed in a slightly different way. Indeed, in Elman and Jordan RNNs the contextual information provided by the recurrent connection is summed to the input information (see equation~\ref{eqn:JordanHiddenContextHistory} above).
In our variants of RNNs instead, word and label embeddings are concatenated and provided to the hidden layer as different inputs.

The most interesting consequence of modifications in our RNN variants, 
is the fact that output labels are mapped into distributional representations, 
as it is usually done for input items.
Indeed, the first layer of our network, is just a mapping from sparse ``one-hot'' representations to distributional representations.
Such mapping results in fine features and attractive syntactic and semantic properties, 
as shown by \textit{word2vec} and similar works \cite{Word2Vec_Mikolov:2013}.
Such representations can be learnt from data the same way as for words.
In the simplest case, this can be done by using sequences of output labels.
When structured information is available, like syntactic parse trees or structured semantic labels such as named entities or entity relations, more sophosticated embeddings can be learnt.
In this work, we learn label embeddings using sequences of output labels associated to word sequences in annotated data.
It is worth noting that the idea of using label embeddings has been introduced by \cite{chen-manning:2014:EMNLP2014} in the context of dependency parsing. 
In this paper, we focus on the use of several label embeddings as context, 
thus encoding label dependencies, which are very important in sequence labeling tasks.

Using the same notation as above,
we name $E_w$ the embedding matrix for words, and $E_l$ the embedding matrix for labels.
We name

{\centering
$I_t = [E_w(v(w_{t-w})) ... E_w(v(w_{t})) ... E_w(v(w_{t+w}))]$

}

the concatenation of the vectors representing the input words when processing the position $t$ in a sequence,
while

{\centering
$L_t = [E_l(v(y_{t-c+1})) E_l(v(y_{t-c+2})) ... E_l(v(y_{t-1}))]$

}

is the concatenation of the vectors representing the output labels predicted at the previous $c$ steps.
The hidden layer activities are computed as:

{\centering
$h_t = \Sigma( [I_t L_t] \cdot H )$

}

$\Sigma$ is the sigmoid activation function \cite{PracticalRecommendations:Bengio:2012}, 
$[\cdot]$ means the concatenation of the two matrices and we omit biases to keep notations lighter.
The remainder of the layer activities, as well as the error computation and back-propagation are computed the same way as in traditional RNNs.

Note that in this variant of RNN there is no $R$ matrix at the recurrent connection.
The recurrent connection here means that the output is given back as input to the network and it is thus converted explicitely from probability distribution given by the softmax into a label index, which is used in turn to select a label embedding from the matrix $E_l$.
Basically the role played from matrix $R$ in Elman and Jordan RNN, is played from matrix $E_l$ in our variant.

Another important interest of having the recursion between output and input layers is robustness.
This is a direct consequence of using embeddings for output labels.
Since we use several predicted labels as context at each position $t$ (see $L_t$ above), 
at least in the later stages of learning (when the model is close to the final optimum),
it is unlikely to have several mistakes in the same context.
Even then, thanks to the properties of distributed representations \cite{LinguisticRegularities:Mikolov:2013},
wrong labels have very similar representations to the correct ones.
Taking an example cited in \cite{LinguisticRegularities:Mikolov:2013}:
if we use \textit{Paris} instead of \textit{Rome}, it has no effect for many NLP tasks,
as they are both proper nouns for POS-tagging, locations for named entity recognition etc.
Distributed representations for labels provide the same robustness on the output side.\footnote{Sometimes in POS-tagging, models mistake verbs and nouns. They make such errors because some particular verbs occur in the same context of nouns (e.g. ``the sleep is important''), and so have similar representations.}

Jordan RNNs cannot provide in general the same robustness.
We can interpret hidden activity computation in a Jordan RNN in two ways.

On the one hand, if we interpret a Jordan RNN as using sparse label representations as input to the hidden layer, such representations are either ``one-hot'' representations of labels or probability distributions given as output by the softmax at the output layer.
In the first case it is clear that a mistake may have more effect than in the second one,
as the only value that is not zero is in a wrong position.
But when probability distributions are used, 
we have found that most of the probability mass is ``picked'' on one or few labels,
which thus does not provide much more softness than a ``one-hot'' representation.
In this interpretation, sparse labels are an additional input for the hidden layer, 
the matrix $R$ on the recurrent connection plays the same role as $H$ does for input words.
The two matrices are then summed to compute the total input of the hidden layer.

On the other hand, we can see the multiplication of sparse representation of labels in equation~\ref{eqn:JordanHiddenContextHistory} as the selection of an embedding for labels from matrix $R$.\footnote{Multiplying a ``one-hot'' representation by a matrix is equivalent to selecting one row of the matrix.}
Even in this interpretation there is a substantial difference between the Jordan RNN and our variant of RNN, \texttt{I-RNN} henceforth ($I$ stands for \textit{Improved}).
In order to understand in detail this difference, we focus on the equations for computing the hidden activities.
For Jordan RNN we have:

{\centering
$h_t = \Sigma( I_t \cdot H + [y_{t-c+1} y_{t-c+2} ... y_{t-1}] \cdot R)$

}

For \texttt{I-RNN} we have:

{\centering
$h_t = \Sigma( [I_t L_t] \cdot H )$

}

where $L_t$ is $[E_l(v(y_{t-c+1})) E_l(v(y_{t-c+2})) ... $
$E_l(v(y_{t-1}))]$, that is the concatenation of $c$ previous label embeddings.

In this second interpretation, in Jordan RNN labels are not directly concerned in the computation of the total input for the hidden layer, since they are not directly multiplied by matrix $H$. Only input context $I_t$ is multiplied by $H$. The result of this multiplication is summed to the result of label embedding selection, performed as $y[t-i] \cdot R, i=1 \dots c-1$. Finally the hidden non-linear function $\Sigma$ is applied.
Mixing input processing $I_t \cdot H$ and label processing $y[t-i] \cdot R$ with a sum, can make sense for the tasks for which the Jordan network was designed \cite{jordan:serial}, as output units were of the same nature as input units (speech signal). However we believe that it doesn't express sufficiently well words and labels interactions in NLP tasks. Also, since $y[t-i] \cdot R$ is an embedding selection for labels, labels and words are not processed in the same way: $I_t$ is already made of word embeddings, which are further transformed by $I_t \cdot H$. This further transformation is not applied to label embeddings.

In \texttt{I-RNN} in contrast, sparse labels are first converted into embeddings with $E_l(v(y_{t-i}))]$, $i=1 \dots c$, and their concatenation results in $L_t$. This matrix is further concatenated to the input context $I_t$.
The result of the concatenation is multiplied by matrix $H$ to compute the total input to the hidden layer.\footnote{We remind the reader that we are omitting biases to keep notation lighter}
Finally the hidden non-linear function $\Sigma$ is applied.
This means that information provided by the input context $I_t$ is not mixed with $L_t$ with a sum like in Jordan RNN. These two data are given to the hidden layer as separated inputs. More in particular, the concatenation of $I_t$ and $L_t$ is performed neuron-wise, that is each hidden neuron receives as input all context words and all context labels, encoding them as a network internal feature. Thus we let the hidden layer itself to learn labels interactions, and words-labels interactions.
This is indeed in agreement with the ``philosophy'' of neural networks, where features designing is turned into features learning.
Since words and labels have different nature in sequence labeling tasks, 
we believe that modeling interactions in this way is more effective.
Also, with the \texttt{I-RNN} architecture, words and labels are processed in the same way, as both are first converted into embeddings, and then ``transformed'' again via multiplication by $H$ matrix.
In order to make results obtained with different RNNs comparable, 
we used the same number of hidden neurons for all RNNs.
In \texttt{I-RNN} thus, each hidden neuron receives as input much more information than Jordan and Elman hidden neurons.

\begin{figure}[t, width=\linewidth]
	\centering
	\footnotesize

	\includegraphics[height=11.0cm]{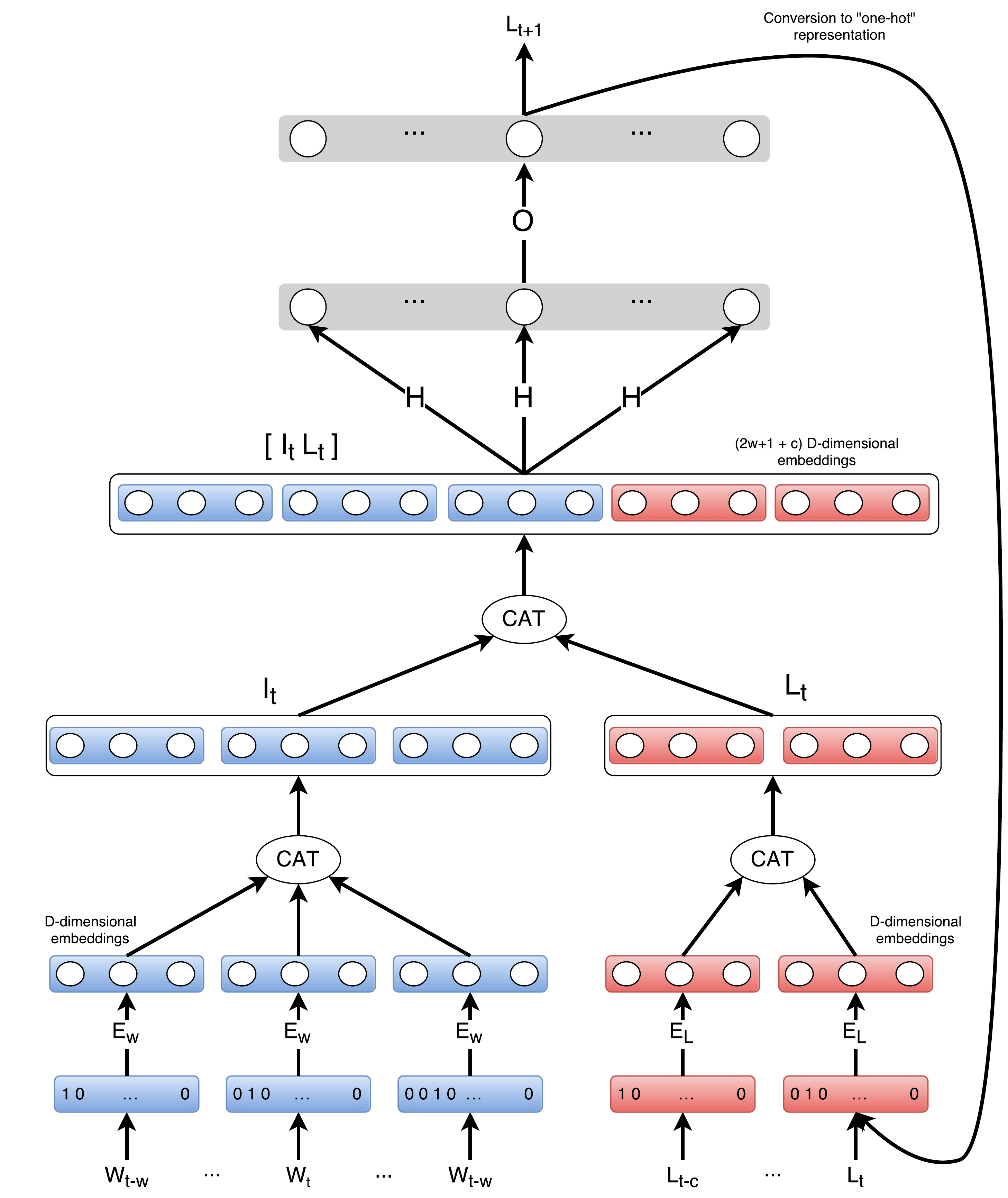}
	\caption{\footnotesize{Detailed architecture of \texttt{I-RNN}. Symbols have the same meaning as equations in the paper, the only exception is that labels are indicated in the figure with upper-case $L$. Please note that matrix $H$ is replicated at the hidden layer computation meaning that all neurons receive the whole $[I_t L_t]$ input, which is made of $2\cdot w + 1 + c$ $D$-dimensional embeddings: $2w+1$ word embeddings and $c$ label embeddings. $CAT$ is the concatenation operator.}}
	\label{fig:IRNN-details}
\end{figure}

In order to make the explanation more clear, 
\texttt{I-RNN} architecture is detailed in figure~\ref{fig:IRNN-details}.
Symbols have the same meaning as equations in the paper, the only exception is that labels are indicated in the figure with upper-case $L$.
Matrix $H$ is replicated at the hidden layer computation meaning that all neurons receive the whole $[I_t L_t]$ input, which is made of $2\cdot w + 1 + c$ $D$-dimensional embeddings: $2w+1$ word embeddings and $c$ label embeddings. $CAT$ is the concatenation operator. Please note that the concatenations of embeddings are performed at two different steps just for the sake of clearness and to be coherent with equations in the paper, 
all concatenations can be performed in one step.

Our second variant of RNN combines the characteristics of an Elman RNN and of our first variant.
In this variant, the only difference with the first one is the computation of the hidden layer activities, where we use the concatenation of the $c$ previous hidden layer states in addition to the information already used in the first variant:

{\centering
$h_t = \Sigma( [I_t L_t] \cdot H + [h_{t-c+1} h_{t-c+2} ... h_{t-1} ] \cdot R )$

}

\subsection{Forward, Backward and Bidirectional RNNs}
\label{subsec:bidir}

All RNNs described in this work are studied in their \textit{forward}, \textit{backward} and bidirectional versions \cite{Schuster:1997:BRNN}.
Forward RNNs work as described so far.
Backward RNNs have exactly the same architecture, 
the only difference being that they process sequences in the reverse order, 
from the end to the begin. Backward RNNs can be used to predict future labels in a sequence.

Bidirectional RNNs use both past and future information to predict the next label, 
which is both words and labels in our variants and in Jordan RNNs,
or both words and hidden layers in Elman RNNs.
When labeling sequences with bidirectional RNNs, 
a backward network is first used to predict labels backward.
The bidirectional RNN then processes sequences in forward direction using past contextual information as usual, 
and future contextual information provided by the states and labels predicted by the backward network.
The hidden layer of the bidirectional version of our fist variant of RNNs is thus computed as:

{\centering
$h_t = \Sigma( [I_t L_t^p L_t^f] \cdot H  )$

}

where $L_t^p = L_t$ introduced above, 
while

{\centering
$L_t^f = [E_l(v(y_{t+1})) \dots E_l(v(y_{t+c-1})) E_l(v(y_{t+c}))]$

}

is the concatenation of the vectors representing the $c$ future labels predicted by the backward model.
It is very similar for our second variant, refer to \cite{Schuster:1997:BRNN} for details.

\subsection{Recurrent Neural Network Complexity}
\label{subsec:RNNComplexity}

We provide here an analysis of the complexity in terms of the number of parameters involved in each model.
In Jordan RNN we count:

{\centering
$|V| \times D + ((2 w + 1) D + c |O|) \times |H| + |H| \times |O|$

}

where $|V|$ is the size of the input dictionary,
$D$ the dimensionality of the embeddings,
$|H|$ and $|O|$ are the size of the hidden and output layers, respectively,
$w$ the size of the window of words used as context on the input side\footnote{The word input context is thus made of $w$ words on the left and $w$ on the right of the word at a given position $t$, plus the word at $t$ itself, which gives a total of $2 w + 1$ input words} and
$c$ is the size of the context of labels, which is multiplied by the dimensionality of the output label dictionary $|O|$.

With the same symbols, in an Elman RNN and in our first variant we have, respectively:

{\centering
$|V| \times D + ((2 w + 1) D + c |H|) \times |H| + |H| \times |O|$

}

and

{\centering
$|V| \times D + |O| \times D + ((2 w + 1) D + c D) \times |H| + |H| \times |O|$

}

The only difference between the Jordan and Elman RNNs lies in the factors $c |O|$ and $c |H|$.
Their difference in complexity depends thus on the size of the output layer (Jordan) with respect to the size of the hidden layer (Elman). Since in sequence labeling tasks the hidden layer is often bigger, Elman RNN is more complex than Jordan RNN.
The difference between the Jordan RNN and our first variant is in the factors $|O| \times D$ and $c D$.
The first is due to the label embeddings\footnote{We use embeddings of the same size $D$ for words and labels.}, the second is due to the use of such embeddings as input to the hidden layer.
Since often $D$ and $O$ have sizes in the same order of magnitude, 
and thanks to the use of vectorized operations on matrices, 
we didn't found a noticeable difference in terms of training and testing time between the Jordan RNN and our first variant.
This simple analysis also shows that our first variant roughly needs the same number of connections in the hidden layer as a Jordan RNN. Our first variant is thus architecturally equivalent to a Jordan RNN.

In contrast, for the second variant we have:

{\centering
$|V| \times D + |O| \times D + ((2 w + 1) D + c D + c |H|) \times |H| + |H| \times |O|$

}

The additional term $c |H|$, is due to the same recurrent connection as in an Elman RNN.
Using vectorized operations for matrix calculations, we found
the second variant slower for both training and testing time by a factor $1.15$ with respect to the other RNNs.
The same complexity analysis holds for backward RNNs. 
But bidirectional RNNs are even more complex. 
Without deriving any new formula, we note that they are slower with respect to their corresponding forward/backward models by a factor of roughly $1.5$.

\section{Evaluation}
\label{sec:eval}

\subsection{Corpora}
\label{subsec:corpora}

We used four distinct corpora:

\textbf{The Air Travel Information System (ATIS)} task \cite{Dahl:1994:ESA:1075812.1075823} has been designed to automatically provide flight information in SLU systems.
The semantic representation is frame-based and the goal is to find the correct frame and the corresponding semantic slots.
For example, for the sentence \textit{``I want the flights from Boston to Philadelphia today''},
the correct frame is \texttt{FLIGHT} and the words \textit{Boston}, \textit{Philadelphia} and \textit{today} must be annotated with the concepts \texttt{DEPARTURE.CITY}, \texttt{ARRIVAL.CITY} and \texttt{DEPARTURE.DATE}, respectively.

ATIS is a relatively simple task dating from $1993$.
The training set is made of $4978$ sentences taken from the ``context independent'' data in the ATIS-2 and ATIS-3 corpora.
The test set is made of $893$ sentences, taken from the ATIS-3 NOV93 and DEC94 datasets. There are no official development data provided with this corpus, we have thus taken a part of the training data at random to play this role. 
\footnote{
Please see \cite{Dahl:1994:ESA:1075812.1075823} for more details on the ATIS corpus.}

\textbf{The French corpus MEDIA} \cite{Bonneau-Maynard2006:media} has been created to evaluate SLU systems providing tourist information, in particular hotel information in France.
It is composed of $1250$ dialogues acquired with a Wizard-of-OZ protocol where $250$ speakers have applied $5$ hotel reservation scenarios.
The dialogues have been annotated following a rich semantic ontology.
Semantic components can be combined to create complex semantic labels.\footnote{For example the label \texttt{localisation} can be combined with \texttt{city}, \texttt{relative-distance}, \texttt{general-relative-place}, \texttt{street} etc. }
In addition to this rich annotation, another difficulty lies in the coreferences.
Some words cannot be correctly annotated without information about previous dialog turns.
For example in the sentence \textit{``Yes, the one at less than fifty euros per night''}, \textit{the one} refers to an hotel previously introduced in the dialog.
Statistics on training, development and test data from this corpus are shown in table~\ref{tab:MEDIAStats}.

Both ATIS and MEDIA can be modelled as sequence labeling tasks using the \textit{BIO} chunking notation \cite{Ramshaw95:BIO}.
Several different works compared on ATIS \cite{RNNforLU:Interspeech:2013,RNNforSLU:Interspeech:2013,Vukotic.etal_2015,Mesnil:RNN:2015}.
\cite{Vukotic.etal_2015} is the only work providing results on MEDIA with RNNs, it also provides results obtained with CRFs \cite{lafferty01:crf}, allowing an interesting comparison.

\begin{table*}[t]
    \centering
    \scriptsize
    \begin{tabular}{|l|rr|rr|rr|}
      \hline
      & \multicolumn{2}{|c|}{training} & \multicolumn{2}{|c|}{development} & \multicolumn{2}{|c|}{test}\\
      \hline
      \# sentences  &\multicolumn{2}{|c|}{12,908} &\multicolumn{2}{|c|}{1,259}&\multicolumn{2}{|c|}{3,005} \\
      \hline
      \hline
      & \multicolumn{1}{|c}{words} & \multicolumn{1}{c|}{concepts} &  \multicolumn{1}{|c}{words} & \multicolumn{1}{c|}{concepts} &
      \multicolumn{1}{|c}{words} & \multicolumn{1}{c|}{concepts} \\
      \hline
      \# words          & 94,466 & 43,078 & 10,849 & 4,705 & 25,606 & 11,383 \\
      \# vocab.         &  2,210 &     99 &    838 &    66 &  1,276 &     78 \\
      \# OOV\%   & --     & --     &  1.33  & 0.02  &  1.39  &  0.04  \\
      \hline
    \end{tabular}
    \caption{Statistics on the corpus MEDIA}
  \label{tab:MEDIAStats}
	\vspace{-1.0em}
\end{table*}

\begin{table}[t]
	\centering
	\scriptsize
	\begin{tabular}{|l|r|r|r|}

	\hline
	Section		& \# sentences	&	\# tokens	&	\# unk. tokens	\\
	\hline
	\hline
	FTB-train		&	9881	&	278083		&						\\
	FTB-dev		&	1235	&	36508		&	1790 (4,9\%)			\\
	FTB-test		&	1235	&	36340		&	1701 (4,7\%)			\\
	\hline

	\end{tabular}
	\caption{Statistics on the FTB corpus used for POS-tagging}
	\label{tab:FTBStats}
	\vspace{-2.0em}
\end{table}

\begin{table*}[t]
    \centering
	\scriptsize
    \begin{tabular}{|l|r|r|r|r|}
      \hline
      Data set & Sections & Sentences & Tokens & Unknown\\
      \hline
	Training		&	0-18	&	38,219	&	912,344	&		0\\
	Development	&	19-21	&	 5,527	&	131,768	&	 4,467\\
	Test			&	22-24	&	 5,462	&	129,654	&	 3,649\\
      \hline
    \end{tabular}
    \caption{Statistics of the training, development and test data texts of the Penn Treebank corpus}
  \label{tab:PennTreebankPOSStats}
\end{table*}

\textbf{The French Treebank (FTB) corpus} is presented in \cite{FTB:Abeille:2003}. 
The version we use for POS-tagging is exactly the same as in \cite{Denis:2012:FTB-POStagging}.
But, in contrast to them, 
who reach the best result on this task with an external lexicon, 
we don't use any external resource here \footnote{\cite{Denis:2012:FTB-POStagging} also provides results without using the external lexicon}.
Statistics on the FTB corpus are shown in table~\ref{tab:FTBStats}.

\textbf{The Penn Treebank (FTB) corpus} is 
presented in \cite{Marcus93buildinga}.
In order to have a direct comparison with previous works \cite{Toutanova03posTagging} \cite{shen-satta-joshi:2007:ACLMain} \cite{Collobert:2011:NLP:1953048.2078186}, 
we split the data as they do: sections $0-18$ are used for training, $19-21$ for validation and $22-24$ for testing. Statistics on the PTB corpus are shown in table~\ref{tab:PennTreebankPOSStats}.

\subsection{RNNs Parameters Settings}
\label{sec:settings}

In order to compare with some published works on the ATIS and MEDIA tasks, 
we use the same dimensionality settings used by \cite{RNNforLU:Interspeech:2013}, \cite{RNNforSLU:Interspeech:2013} and \cite{Vukotic.etal_2015},
that is embeddings have $200$ dimensions, hidden layer has $100$ dimensions.
We also use the same context size for words, that is $w = 3$, and we use $c = 6$ as labels context size in our variants and in Jordan RNN.
We use the same tokenization, basically consisting of words lower-casing.

In contrast, in our models the sigmoid activation function at the hidden layer and the $L2$ regularization.
While \cite{RNNforLU:Interspeech:2013}, \cite{RNNforSLU:Interspeech:2013},
\cite{Vukotic.etal_2015} and \cite{Mesnil:RNN:2015} use the rectified linear activation function and the dropout regularization \cite{PracticalRecommendations:Bengio:2012} \cite{JMLR:v15:srivastava14a}.

For the FTB POS-tagging task we have used $200$-dimensional embeddings, $300$-dimensional hidden layer, again $w = 3$ for the context on the input side,
and $6$ context labels on the output side. The bigger hidden layer gave better results during validation experiments, due to the larger word dictionary in this task with respect to the others, 
roughly $25000$ for the FTB against $2000$ for MEDIA and $1300$ for ATIS.
In contrast to \cite{Denis:2012:FTB-POStagging}, 
which has used several features of words (prefixes, suffixes, capitalisation information etc.), we only performed a simple tokenization for reducing the size of the input dictionary: all numbers have been mapped to a conventional symbol (NUM), and nouns not corresponding to proper names and starting with a capital letter have been converted to lowercase.
We preferred this simple tokenisation without using rich features, because our goal in this work is not obtaining the best results ever,
it is to compare Jordan and Elman RNNs with our variants of RNN and show that our variants works better for sequence labeling.
Adding many features and/or building sophisticated models would make the message less clear,
as results would be probably better but the improvements could be attributed to rich and sophisticated models, instead of to the model itself.

For the PTB POS-tagging task we use exactly the same settings and pre-processing as for the FTB task, 
except that we used $400$ hidden neurons.
During validation we found that this works better, 
again due to the size of the dictionary which is $45000$ for this task (after pre-processing).

We trained all RNNs with exactly the same protocol:
i) we first train neural language models to obtain word and label embeddings.
This language model is like the one in \cite{Bengio03aneural}, except it uses both words/labels in the past and in the future to predict next word/label.
ii) we train all RNNs using the same embeddings trained at previous step.
We train the RNN for word embeddings for $20$ epochs, the RNN for label embeddings for $10$ epochs, and we train the RNNs for sequence labeling for $20$ epochs.
The number of epochs has been roughly optimized on development data.
At the end of training we keep the model which gave the best tagging accuracy on development data.
Also we roughly optimized on development data the learning rate and the parameter $\lambda$ for regularization, the best values found are $0.5$ and $0.003$, respectively.

\subsection{Training and Tagging Time}
\label{subsec:time}

Since implementations of RNNs use in this work are prototypes\footnote{Our implementations are basically written in Octave https://www.gnu.org/software/octave/}, it does not make sense to compare them to state-of-the-art in terms of training and tagging time.
However it is worth providing training times at least to have an idea and to have a comparison among different RNNs.

As explained in section~\ref{subsec:RNNComplexity}, our first variant \texttt{I-RNN} and the Jordan RNN have the same complexity. Also, since the size of the hidden layer is in the same order of magnitude as the size of the output layer (i.e. the number of labels), also Elman RNN has roughly the same complexity as the Jordan RNN. This is reflected in the training time.

The training time for label embeddings is always relatively short, 
as the size of the output layer, that is the number of labels, is always relatively small.
This training time thus can vary from few minutes for ATIS and MEDIA, to less that $1$ hour for the FTB corpus.

Training word embeddings is also very fast on ATIS and MEDIA, taking less than $30$ minutes for ATIS and roughly $40$ minutes for MEDIA.
In contrast training word embeddings on the FTB corpus takes roughly $5$ days, training time goes to roughly $6$ days for the PTB word embeddings.

Training the RNN taggers takes roughly the same time as training the word embeddings, 
as the size of the word dictionary is the dimension that most affects the computational complexity at softmax used to predict next label.

Concerning the second variant of RNN proposed in this work, \texttt{I+E-RNN}, this is slower as it is more complex in terms of number of parameters. Training \texttt{I+E-RNN} on ATIS and MEDIA takes roughly $45$ minutes and $1$ hour, respectively. In contrast training \texttt{I+E-RNN} on the FTB and PTB corpora takes roughly $6$ days and $7$ days, respectively.

We didn't keep track of tagging time, however it is always negligible with respect to training time, and it is always measured in few minutes.
All times provided here are with a $1.7$ GHz CPU, single process.

\subsection{Sequence Labeling Results}
\label{subsec:results}

The evaluation of all models described here is shown in table~\ref{tab:SLUATIS} \ref{tab:SLUMEDIA} \ref{tab:POSFTB} and \ref{tab:POSPTB}, in terms of F1 measure for ATIS and MEDIA, \textit{Accuracy} on the FTB and PTB.
Our implementations of Elman and Jordan RNN are indicated in tables as \texttt{E-RNN} and \texttt{J-RNN}.
Our new variants are indicated as \texttt{I-RNN} and \texttt{I+E-RNN}.

Table~\ref{tab:SLUATIS} shows results on the ATIS corpus.
In the higher part of the table we show results obtained using only words as input.
In the lower part, results are obtained using both words and word-classes available for this task.
Such classes concern city names, airports and time expressions.
They allow the models to generalize from specific words triggering concepts.\footnote{For example the cities of \textit{Boston} and \textit{Philadelphia} in the example above are mapped to the class \texttt{CITY-NAME}. If a model has never seen \textit{Boston} during the training phase, but has seen at least one city name, it will possibly annotate \textit{Boston} as a departure city thanks to some discriminative context, such as the preposition \textit{from}.}

Note that our results on the ATIS corpus are not always comparable with those published in the literature because:
\textit{i)} Models published in the literature use a \textit{rectified linear} activation function at the hidden layer\footnote{$f(x) = max(0,x).$} and the \textit{dropout} regularization.
Our models use the sigmoid activation function and the $L2$ regularization.
\textit{ii)} For experiments on ATIS we have used roughly $18\%$ of the training data for development, we thus used a smaller training set.
\textit{iii)} The works we compare with do not always give details on how classes available for the task have been integrated into their models.
\textit{iv)} Layer dimensionality and hyper-parameter settings do not always match those of published works. In fact, to avoid running too much experiments, we have based our settings on known works, but this doesn't allow a straightforward comparison with other published works.

Despite this, the message we want to claim in this work is still true for two reasons:
\textit{i)} Some of the results obtained with Elman and Jordan RNNs are close, or even better, than state-of-the-art. Thus they are not weak baselines.
\textit{ii)} We provide a fair comparison of our variants with traditional Elman and Jordan RNNs.

The results in the higher part of table~\ref{tab:SLUATIS} show that the best model on the ATIS task, with these settings, is the Elman RNN of \cite{Mesnil:RNN:2015}.
Note that it is not clear how the improvements of \cite{Mesnil:RNN:2015} with respect to \cite{RNNforSLU:Interspeech:2013} (in part due to same authors) have been obtained.
Indeed, in \cite{RNNforSLU:Interspeech:2013} authors obtain the best result with a Jordan RNN, 
while in \cite{Mesnil:RNN:2015} an Elman RNN gets the best performance.
During our experiments,
using the same experimentation protocol as \cite{Mesnil:RNN:2015}, 
we could not reach the same performances.
We conclude that the difference between our results and those in \cite{Mesnil:RNN:2015} are due to reasons mentioned above.
Beyond this, we note that our Elman and Jordan RNN implementations are equivalent to those of \cite{RNNforSLU:Interspeech:2013}. Also, our first variant of RNNs, \texttt{I-RNN}, obtains the second best result ($93.84$ in bold), which is the best result we could reach. Our second variant is roughly equivalent to a Jordan RNN on this task.

The results in the lower part of the table~\ref{tab:SLUATIS} (\textit{Classes}), obtained with both words and classes as input, are quite better than those obtained with words only.
They roughly follow the same behavior, except that in this case our Jordan RNN is slightly better than our second variant. The first variant \texttt{I-RNN} obtains again the best result among our implementations ($95.21$ in bold).
In this case also, we attribute the differences with respect to published results to the different settings mentioned above.
For comparison, in this part of the table we show also results obtained using CRF.

On the ATIS task, using either words or both words and classes as inputs,
we can see that results are always quite high.
This task is relatively simple.
Label dependencies can be easily modeled, as there is basically no segmentation of concepts over different consecutive words (one concept corresponds to one word).
In this settings,
the potential of our new variants of RNNs cannot be fully exploited.
This limitation is confirmed by the results obtained by \cite{Vukotic.etal_2015} and \cite{Mesnil:RNN:2015} using CRFs.
Indeed, RNNs don't take the whole sequence of labels into account in their decision function.
In contrast, CRFs use a global decision function taking all the possible labelings of a given input sequence into account to predict the best sequence of labels.
The fact that CRFs are less effective than RNN on ATIS
is a clear sign that label dependency modeling is relatively simple in this task.

Table~\ref{tab:SLUMEDIA} shows the results on the corpus MEDIA.
As already mentioned, this task is more difficult because of its richer semantic annotation, 
and also because of the coreferences and of the segmentation of concepts over several words.
This creates relatively long label dependencies, 
which cannot be taken into account by simple models.
The difficulty of this task is confirmed by the magnitude of results ($10$ F1 points lower than on the ATIS task).

As can be seen in Table~\ref{tab:SLUMEDIA}, 
CRFs are in general much more effective than Jordan and Elman RNNs on MEDIA.
This outcome could be expected as RNNs use a local decision function not able to take long label dependencies into account.
We can also see that our implementations of Elman and Jordan RNNs are comparable, even better in the case of Elman RNN, with state-of-the-art RNNs of \cite{Vukotic.etal_2015}. 

More importantly, results on the MEDIA task shows that in this particular experimental settings where taking label dependencies into account is crucial, our new variants are remarkably more effective than both our implementations and state-of-the-art implementations \cite{Vukotic.etal_2015} of Elman and Jordan RNNs.
This holds for forward, backward and bidirectional RNNs.
Moreover, the bidirectional version of our variants of RNNs outperforms even CRF.
We attribute this effectiveness to a better modeling of label dependencies, due to label embeddings.

Table~\ref{tab:POSFTB} shows the results obtained on the POS-tagging task of the FTB corpus.
On this task, we compare our RNN implementations to the state-of-the-art results obtained in \cite{Denis:2012:FTB-POStagging} with the model $MElt_{fr}^0$.
We would like to underline that $MElt_{fr}^0$, when it does not use external resources like in the model obtaining the best absolute result in \cite{Denis:2012:FTB-POStagging}, nevertheless uses several features associated with words that provide an advantage over features used in our RNNs.
As can be expected thus, $MElt_{fr}^0$ outperforms the RNNs.
Results in table~\ref{tab:POSFTB} shows that forward and backward RNNs are quite close to each other on this task, \texttt{I-RNN}, providing a little improvement.
In contrast, the bidirectional version of \texttt{I-RNN} and \texttt{I+E-RNN} provide a significant improvement over Jordan and Elman RNNs.

Table~\ref{tab:POSPTB} shows the results obtained on the WSJ POS tagging task.
We also provide the results of \cite{Toutanova03posTagging} and \cite{shen-satta-joshi:2007:ACLMain} for comparison with the state-of-the-art.
This is just for a matter of comparison, as the results shown in those works were achieved with rich features and more sophisticated models.
Instead, we can roughly compare our results with those of \cite{Collobert:2011:NLP:1953048.2078186}, 
which were also obtained with neural networks. In particular, we can compare with the model called \texttt{NN+SLL}.
Note that this is a rough comparison as the model of \cite{Collobert:2011:NLP:1953048.2078186}, 
though not a RNN, integrates capitalisation features and uses a convolution and a \textit{max-over-time} layer to encode large context information.

Results for POS tagging on the PTB corpus roughly confirm the conclusions reached for FTB results, 
with the only difference that in this case \texttt{I+E-RNN} is slightly better than \texttt{I-RNN}.
Our RNNs don't improve the state-of-the-art, but are all more effective than the model of \cite{Collobert:2011:NLP:1953048.2078186}.
This result is particularly important, as it shows that RNNs, even without using a sophisticated encoding of the context like the model \texttt{NN+SLL} in \cite{Collobert:2011:NLP:1953048.2078186}, 
are intrinsecally a better model for sequence labeling. This claim is enforced also by the fact that \texttt{NN+SLL} of \cite{Collobert:2011:NLP:1953048.2078186} implements a global probability computation strategy similar to CRF (SLL stands for \textit{Sentence Level Likelihood}), while all RNNs presented here use a local decision function (see equation~\ref{eq:dec-function}).
Again, the ranking of RNN models on the PTB POS-tagging task is stable, the variants of RNNs proposed in this work being more effective than traditional Elman and Jordan RNNs.

We would like to notify that we have also performed some experiments on ATIS and MEDIA without pre-training label embeddings. The results reached are not substantially different from those obtained with pre-training label embeddings. Indeed, on relatively small data, it is not rare to have similar or even better results without pre-training.
This is due to the fact that learning effective embeddings requires a relatively large amount of data.
\cite{Vukotic.etal_2015} also shows results obtained with embeddings pre-trained with \textit{word2vec} and without embedding pre-training. His conclusions are indeed very similar.
More generally, reaching roughly the same results on a difficult task like MEDIA without label embedding pre-training 
is a clear sign that our variants of RNNs are superior to traditional RNNs because they use a context made of label embeddings. As a matter of fact, the gain with respect to Elman and Jordan RNNs cannot be attributed in this case to the use of pre-trained embeddings.
On relatively larger corpora like FTB and PTB, label embedding pre-training seems to provide a slight improvement.

Finally, we have run experiments also modifying Jordan and Elman RNNs so that to model words and labels interactions more like \texttt{I-RNN} does, that is word and label embeddings (or hidden states in Elman RNN) are not summed together, instead they are concatenated.
Results obtained were not substantially different from those obtained with ``traditional'' Jordan and Elman RNN and in any case \texttt{I-RNN} was always performing best, still keeping a large gain over Elman and Jordan RNN on the MEDIA task.
The explanation of this outcome is that keeping word and label embeddings separated, and then multiplying both by matrix $H$ to compute hidden activities, as we do in I-RNN, is more effective than concatenating $I_t \cdot H$ and $y[t-1] \cdot R$, as we did for Jordan RNN, and analogously with previous hidden layer for Elman RNN.
This is not surprising, as \texttt{I-RNN} also in this case is applying one transformation more when multiplying label embeddings by $H$ to compute the total input to the hidden layer.

It is someway surprising that \texttt{I-RNN} systematically outperforms \texttt{I+E-RNN}, the latter model integrates more information at the hidden layer and thus should be able to take advantage of both Elman and \texttt{I-RNN} characteristics. While an analysis to explain this outcome is not trivial, 
our interpretation is that using two recursions in a RNN gives actually redundant information to the model.
Indeed, the output of the hidden layer keeps the internal state of the network, which is the internal (distributed) representation of the input n-gram of words around position $t$ and the previous $c$ labels.
The recursion at the hidden layer allows to keep this information ``in memory'' and to use it at the next step $t+1$. However using the recursion of \texttt{I-RNN} the previous $c$ labels are also given explicitely as input to the hidden layer. This may be redundant, and constrain the model to learn an increased amount of noise.
A similar idea of hybrid RNN model has been tested in \cite{Mesnil:RNN:2015} without showing a clear advantage on Elman and Jordan RNNs.

What can be said in general from the results obtained on all the presented tasks, 
is that RNN architectures using label embedding context can model label dependencies in a more effective way, even when these dependencies are relatively simple (like in ATIS and POS tagging tasks).
The two variant of RNNs proposed in this work, in particular the \texttt{I-RNN} variant, 
are for this reason more effective than Elman and Jordan RNNs on sequence labeling tasks.

\begin{table}[t]
    \centering
    \scriptsize
    \begin{tabular}{|l|r|r|r|}
      \hline
      Model & \multicolumn{3}{c|}{F1 measure} \\
        \hline
	& \textit{forward} & \textit{backward} & bidirectional \\
        \hline
	\hline
	\multicolumn{4}{|c|}{Words} \\
	\hline
	\cite{RNNforSLU:Interspeech:2013} E-RNN	& 93.65\% & 92.12\% & -- \\
	\cite{RNNforSLU:Interspeech:2013} J-RNN	& 93.77\% & 93.31\% & 93.98\% \\
	\cite{Mesnil:RNN:2015} E-RNN			& \textbf{94.98\%} & -- & 94.73\% \\
	\texttt{E-RNN}							& 93.41\% & 93.18\% & 93.41\% \\
	\texttt{J-RNN}							& 93.61\% & 93.55\% & 93.74\% \\
	\texttt{I-RNN}							& \textbf{93.84\%} & 93.56\% & 93.73\% \\
	\texttt{I+E-RNN}						& 93.74\% & 93.44\% & 93.60\% \\
	\hline
	\hline
	\multicolumn{4}{|c|}{Classes} \\
	\hline
	\cite{Vukotic.etal_2015} E-RNN			& 96.16\% & -- & -- \\
	\cite{Vukotic.etal_2015} CRF				& -- & -- & 95.23\% \\
	\cite{RNNforLU:Interspeech:2013} E-RNN	& 96.04\% & -- & -- \\
	\cite{Mesnil:RNN:2015} E-RNN			& 96.24\% & -- & \textbf{96.29\%} \\
	\cite{Mesnil:RNN:2015} CRF				& -- & -- & 95.16\% \\
	\texttt{E-RNN}							& 94.73\% & 93.61\% & 94.71\% \\
	\texttt{J-RNN}							& 94.94\% & 94.80\% & 94.89\% \\
	\texttt{I-RNN}							& \textbf{95.21\%} & 94.64\% & 94.75\% \\
	\texttt{I+E-RNN}						& 94.84\% & 94.61\% & 94.79\% \\
	\hline
    \end{tabular}
    \caption{Results of SLU on the ATIS corpus}
  \label{tab:SLUATIS}
\end{table}

\begin{table}[t]
    \centering
    \scriptsize
    \begin{tabular}{|l|r|r|r|}
      \hline
      Model & \multicolumn{3}{c|}{F1 measure} \\
        \hline
        \hline
	& \textit{forward} & \textit{backward} & bidirectional \\
	\hline
	\cite{Vukotic.etal_2015} E-RNN	& 81.94\% & -- & -- \\
	\cite{Vukotic.etal_2015} J-RNN	& 83.25\% & -- & -- \\
	\cite{Vukotic.etal_2015} CRF		& -- & -- & \textbf{86.00\%} \\
	\texttt{E-RNN}					& 82.64\% & 82.61\% & 83.13\% \\
	\texttt{J-RNN}					& 83.06\% & 83.74\% & 84.29\% \\
	\texttt{I-RNN}					& 84.91\% & 86.28\% & \textbf{86.71\%} \\
	\texttt{I+E-RNN}				& 84.58\% & 85.84\% & 86.21\% \\
      \hline
    \end{tabular}
    \caption{Results of SLU on the MEDIA corpus}
  \label{tab:SLUMEDIA}
\end{table}

\begin{table}[t]
    \centering
    \scriptsize
    \begin{tabular}{|l|r|r|r|}
      \hline
      Model & \multicolumn{3}{c|}{Accuracy} \\
        \hline
        \hline
	& \textit{forward} & \textit{backward} & bidirectional \\
	\hline
	\cite{Denis:2012:FTB-POStagging} $MElt_{fr}^0$	& -- & -- & \textbf{97.00\%} \\
	\texttt{E-RNN}								& 96.31\% & 96.30\% & 96.32\% \\
	\texttt{J-RNN}								& 96.31\% & 96.27\% & 96.32\% \\
	\texttt{I-RNN}								& 96.36\% & 96.37\% & \textbf{96.51\%} \\
	\texttt{I+E-RNN}							& 96.28\% & 96.33\% & 96.48\% \\
      \hline
    \end{tabular}
    \caption{Results of POS-tagging on the FTB}
  \label{tab:POSFTB}
\end{table}

\begin{table}[t]
    \centering
    \scriptsize
    \begin{tabular}{|l|r|r|r|}
      \hline
      Model & \multicolumn{3}{c|}{Accuracy} \\
        \hline
        \hline
	& \textit{forward} & \textit{backward} & bidirectional \\
	\hline
	\cite{Toutanova03posTagging}				& -- & -- & 97.24\% \\
        \cite{shen-satta-joshi:2007:ACLMain}			& -- & -- & \textbf{97.33\%} \\
        \cite{Collobert:2011:NLP:1953048.2078186}	NN+SLL	& -- & -- & 96.37\% \\
	\texttt{E-RNN}								& 96.75\% & 96.76\% & 96.75\% \\
	\texttt{J-RNN}								& 96.71\% & 96.69\% & 96.77\% \\
	\texttt{I-RNN}								& 96.75\% & 96.72\% & 96.90\% \\
	\texttt{I+E-RNN}							& 96.73\% & 96.71\% & \textbf{96.93\%} \\
      \hline
    \end{tabular}
    \caption{Results of POS-tagging on the PTB}
  \label{tab:POSPTB}
\end{table}

\subsection{Comparison of Jordan-RNN and \texttt{I-RNN} label representations}
\label{subsec:LabelComparison}

We compare Jordan RNN and \texttt{I-RNN} label representations under the interpretation where Jordan RNN hidden activity computation uses sparse labels as an additional input to the hidden layer, as explained in section~\ref{subsec:NewRNN}.
As explained also in the same section, under the other interpretation \texttt{I-RNN} provides the advantage of performing an additional transformation on labels, and gives words and labels embeddings as separated inputs to the hidden layer.

Under the first interpretation, the advantage of using label embeddings in \texttt{I-RNN} instead of ``one-hot'' or probability distribution representations like in Jordan RNNs, is an increased amount of \textit{signal} flowing across the network.
A semantic interpretation of the interaction of these two representations with the network is not trivial.
Indeed, in the probability representation output by the softmax in a Jordan RNN, 
the different dimensions are just probabilities associated to different labels.
In contrast, in label embeddings used in \texttt{I-RNN}, the dimensions are different distributional features, 
related to how a particular label is used in particular label contexts.
A comparison between these two representation is thus not really meaningful.

Instead, we performed a simple analysis of the magnitude of values found in the probability distribution used as representations in Jordan RNNs, when using development data of the MEDIA corpus.
We summarize this analysis as follows:

\begin{enumerate}
\item $7843$ out of $11051$ (~$71$\%) of the time, the maximum value is greater than $0.9$ 
\item $9928$ out of $11051$ (~$90$\%) of the time, the sum of the $3$ highest probabilities is greater than $0.9$
\item Excluding the $3$ highest probabilities, the remaining values in the distribution have very small values (less than $0.001$)
\end{enumerate}

This simple analysis shows that the probability distributions used for label representations in Jordan RNNs do not provide much more information to the network than a ``one-hot'' representation, and not much signal into the network. This problem is someway similar to the ``vanishing gradient'' problem \cite{Hochreiter:GradientVanishing:2001}: as the network learns, the probability gets concentrated on few dimensions and all the other values get very small, limiting network learning.
This problem is all the more obvious that label dependency modeling is more important for the task.
On an absolute scale however, this is less serious than the vanishing gradient problem, as Jordan RNNs still reach competitive performances.

\section{Conclusions}
\label{sec:conclusions}

In this paper we have studied different architectures of Recurrent Neural Networks for sequence labeling tasks.
We have proposed two new variants of RNNs to better model label dependencies, 
and we have compared these variants to the traditional architectures of Elman and Jordan RNNs.
We have explained the advantages provided by the proposed variants with respect to previous RNNs.
We have evaluated all RNNs, either new or traditional, on $4$ different tasks: two of Spoken Language Understanding and two of POS-tagging.
The results show that, even though RNNs don't always improve the state-of-the-art, 
our new variants of RNNs always outperform the traditional Elman and Jordan RNNs.


\bibliographystyle{splncs}
\bibliography{2016_arXiv_NewRNN_author-final}
\end{document}